\documentclass[conference]{IEEEtran}
\IEEEoverridecommandlockouts
\usepackage{cite}
\usepackage{amsmath,amssymb,amsfonts}
\usepackage{graphicx}
\usepackage{textcomp}
\usepackage{xcolor}
\usepackage{url}
\usepackage{subcaption}
\usepackage{booktabs}   
\usepackage{multirow}   
\usepackage{algorithm}
\usepackage{algorithmicx}
\usepackage{algpseudocode}

\usepackage{float} 

\newcommand{\etal}{\textit{et al.}}

\def\BibTeX{{\rm B\kern-.05em{\sc i\kern-.025em b}\kern-.08em
    T\kern-.1667em\lower.7ex\hbox{E}\kern-.125emX}}

\begin{document}

\title{Variance-Reduction Guidance: Sampling Trajectory Optimization for Diffusion Models}


\author{\IEEEauthorblockN{Shifeng Xu}
  \IEEEauthorblockA{\small \textit{College of Computing and Data Science,} \\
    \textit{Nanyang Technological University} \\
    \textit{Singapore} \\
    shifeng001@e.ntu.edu.sg} \and
  \IEEEauthorblockN{Yanzhu Liu}
  \IEEEauthorblockA{\small \textit{Institute for Infocomm Research (I2R) \&}\\
    \textit{Centre for Frontier AI Research, A*STAR}\\
    \textit{Singapore} \\
    liu\_yanzhu@i2r.a-star.edu.sg} \and
  \IEEEauthorblockN{Adams Wai-Kin Kong}
  \IEEEauthorblockA{\small \textit{College of Computing and Data Science,} \\
    \textit{Nanyang Technological University} \\
    \textit{Singapore} \\
    adamskong@ntu.edu.sg}
}

\maketitle

\begin{abstract}
Diffusion models have become emerging generative models. 
Their sampling process involves multiple steps, 
and in each step the models predict the noise from a noisy sample. 
When the models make prediction, the output deviates from the ground truth, 
and we call such a deviation as \textit{prediction error}. 
The prediction error accumulates over the sampling process and deteriorates generation quality. 
This paper introduces a novel technique for statistically measuring the prediction error 
and proposes the Variance-Reduction Guidance (VRG) method to mitigate this error. 
VRG does not require model fine-tuning or modification. 
Given a predefined sampling trajectory, it searches for a new trajectory 
which has the same number of sampling steps but produces higher quality results.
VRG is applicable to both conditional and unconditional generation. 
Experiments on various datasets and baselines demonstrate that 
VRG can significantly improve the generation quality of diffusion models. 
Source code is available at \url{https://github.com/shifengxu/VRG}. 
\end{abstract}


\section{Introduction}
Diffusion models \cite{nips_ddpm,iclr_ddim,lu2022dpm} are emerging and have been used to generate various contents, 
including images \cite{stable_diffusion,diff_autoencoder}, video \cite{videoDM}, 
and others \cite{oord2016wavenet,brown2020language,url_dalle2,cvpr_tcig}. 
The foundational works, Denoising Diffusion Probabilistic Models (DDPM) \cite{nips_ddpm} 
and Denoising Diffusion Implicit Models (DDIM) \cite{iclr_ddim}, 
use 1000-step sampling processes, which are too slow for real-time applications. 
To speed up the sampling process, various approaches \cite{song2020score,lu2022dpm,zhang2022deis,liu2022pndm,wiz2023splitnm} 
have been proposed. 
However, these approaches mainly focus on the accuracy of individual sampling steps, 
overlooking the holistic connection between them. 
Between two adjacent steps, diffusion models not only transform samples but also transfer error. 
While the former is intended, the latter hinders the generation quality. 
Specifically, in each step, the model predicts the noise from noisy sample. 
However, the predicted noise  deviates from the ground-truth noise; 
we term this deviation as \textit{prediction error}. 
Prediction error exists in every sampling step, and accumulates across sampling process. 
\begin{figure*}[!h]
\centering
  \subcaptionbox{Sampling trajectories} 
    {\includegraphics[width=.315\textwidth]{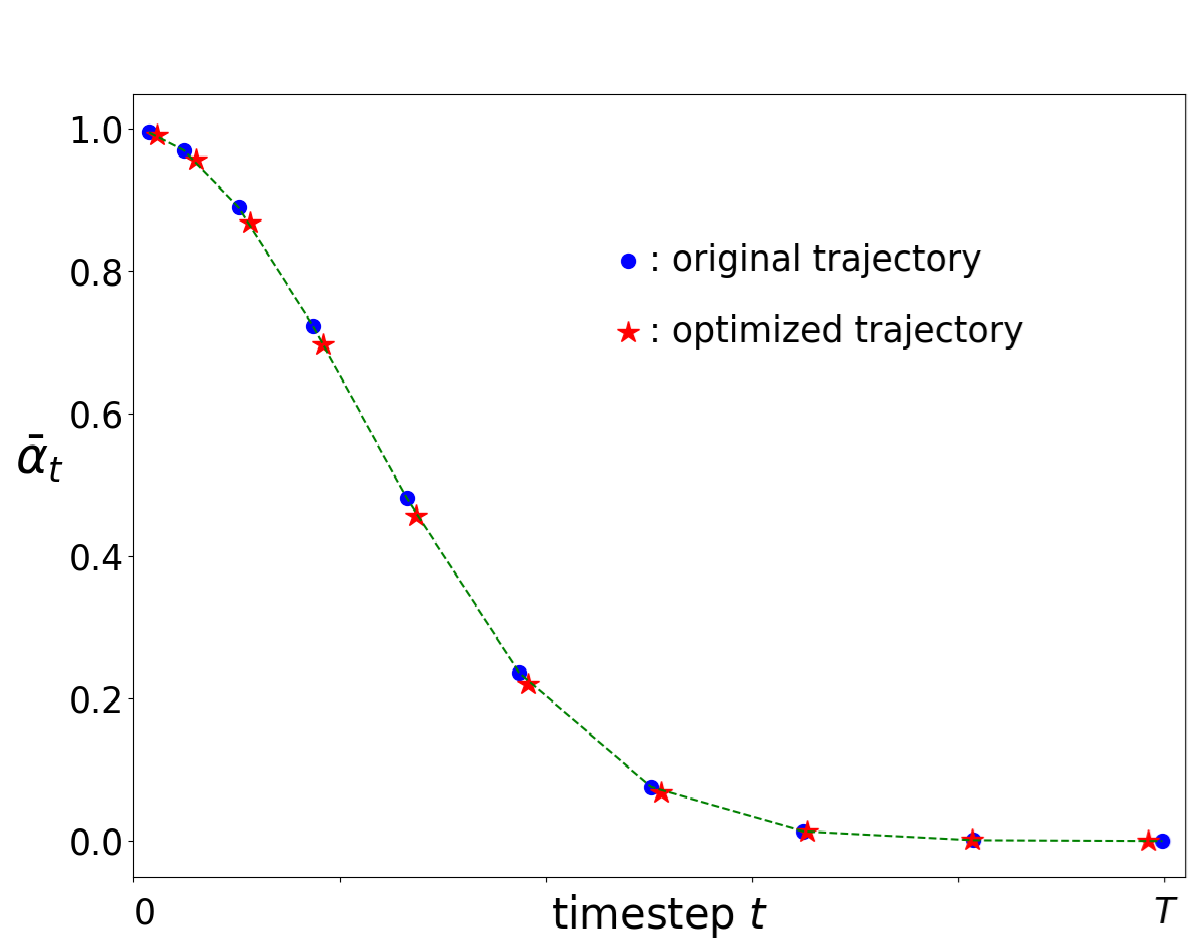}}
  \subcaptionbox{Image generation progress by original (\textit{top}) and optimized (\textit{bottom}) trajectories}
    {\includegraphics[width=.67\textwidth]{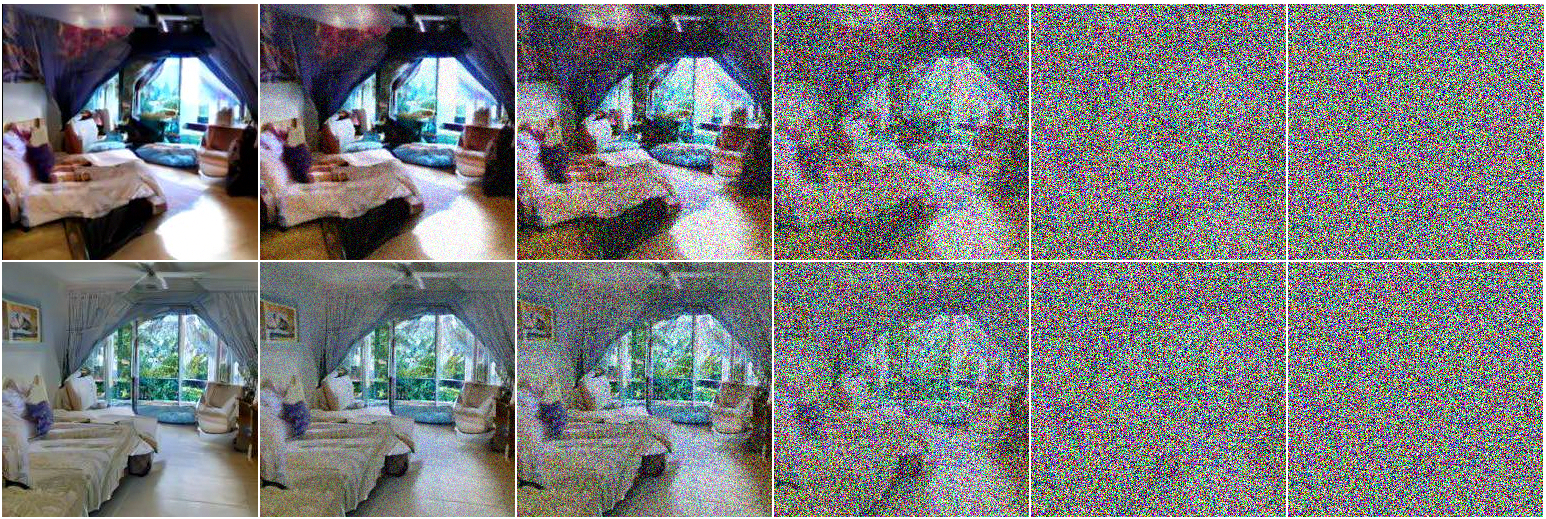}}
\caption{Sampling process of DPM-Solver with order=$2$, schedule=$quadratic$ and steps=$10$. 
         (a) Original and optimized trajectories. 
         The former has noise level sequence $\{\bar{\alpha}_k\}_{k=1}^K$ as 
         \{0.995, 0.97, 0.88, 0.72, 0.48, 0.23, 0.075, 0.0137, 0.00118, 0.000040\}, and the latter has 
         \{0.991, 0.95, 0.86, 0.69, 0.45, 0.22, 0.068, 0.0129, 0.00121, 0.000054\}. 
         (b) Column 1 is generated images, column 6 is the initial Gaussian noises, 
         and columns 2 - 5 are the intermediate results of the sampling progress. 
        } 
\label{fig:compare-ab-img}
\end{figure*}

This paper derives that the cumulative prediction error is the sum of weighted prediction error 
from each sampling step, where the weights are determined by sampling trajectory. 
Sampling trajectory is the sequence of noise levels by which the models convert noise into sample. 
Fig. \ref{fig:compare-ab-img}(a) provides examples of such sequences. 
We propose the \textit{Variance-Reduction Guidance} (VRG) method 
to mitigate the cumulative prediction error by optimizing the sampling trajectory. 
Given an existing trajectory, VRG searches for a new trajectory by reducing the variance of cumulative prediction error. 
The optimized trajectory retains the same number of sampling steps, but enhances generation quality. 
Fig. \ref{fig:compare-ab-img}(b) shows images generated by original and optimized trajectories. 
The optimized trajectory yields more realistic results, demonstrating the benefits of the VRG method. 
VRG is training-free and model-agnostic. 
It operates effectively in both continuous-time and discrete-time scenarios 
and can be seamlessly applied to conditional as well as unconditional generation tasks.

\section{Preliminaries}

\subsection{Diffusion Models}

In diffusion models, $x_{0}\in \mathbb{R}^d$ denotes a $d$-dimensional clean sample from data distribution $q(x_0)$, 
$t\in[0,T]$ indicates the timestep where $x_T$ is pure Gaussian noise. 
The noisy sample $x_{t}$ is obtained by injecting Gaussian noise into $x_0$ : 
\begin{equation}\label{eq:ddpm-x0}
x_t=\sqrt{\bar{\alpha}_t}x_0+\sqrt{1-\bar{\alpha}_t}\epsilon^{(t)} , \epsilon^{(t)}\sim\mathcal{N}(\mathbf{0},\mathbf{I}),
\end{equation}
where $\bar{\alpha}_t\in[0,1]$ controls the noise level and $\mathbf{I}$ is an identity matrix. 
In the sampling process, diffusion models remove noise gradually from $x_T$ and end with a new sample $\tilde{x}_0$. 
Specifically, a neural network $\theta$ is utilized to model the function $\epsilon_{\theta}(x_t, t)$, 
which takes noisy sample $x_t$ and timestep $t$ as inputs, and predicts the noise in $x_t$. 
We denote $\epsilon_{\theta}^{(t)}$ as the value of $\epsilon_{\theta}(x_t,t)$, 
differing with the ground-truth noise $\epsilon^{(t)}$. 

\subsection{Denoising Diffusion Implicit Model}
The sampling process of DDIM \cite{iclr_ddim} follows \eqref{eq:ddim-sampling}, 
which is deterministic in each step: 
\begin{equation}\label{eq:ddim-sampling}
\begin{split}
x_{t-1}&=\sqrt{\bar{\alpha}_{t-1}}\ 
        \frac{x_t-\sqrt{1-\bar{\alpha}_t}\ \epsilon^{(t)}}{\sqrt{\bar{\alpha}_t}} + 
        \sqrt{1-\bar{\alpha}_{t-1}}\ \epsilon^{(t)} \\
       &=\frac{1}{\sqrt{\alpha_t}}\ x_t -
	\frac{\sqrt{1-\bar{\alpha}_t}-\sqrt{\alpha_t-\bar{\alpha}_t}}{\sqrt{\alpha_t}} \epsilon^{(t)} .
\end{split}
\end{equation}
By substituting predicted noise $\epsilon_{\theta}^{(t)}$ into ground-truth noise $\epsilon^{(t)}$, 
we obtain the formula of the predicted sample $\tilde{x}_{t-1}$ in \eqref{eq:ddim-sampling-pred}. 
This is the basis of our proposed method. 
\begin{equation}\label{eq:ddim-sampling-pred}
\tilde{x}_{t-1}=\frac{1}{\sqrt{\alpha_t}} x_t -
	\frac{\sqrt{1-\bar{\alpha}_t}-\sqrt{\alpha_t-\bar{\alpha}_t}}{\sqrt{\alpha_t}} \epsilon_{\theta}^{(t)} . \\
\end{equation}

\subsection{Sampling Trajectory}\label{sec:trajectory}
In \eqref{eq:ddim-sampling} and \eqref{eq:ddim-sampling-pred}, 
the current timestep is $t$ and next timestep is $t-1$, where step size is $1$. 
To expedite sampling, step size may grow larger than 1 and 
next step $s$ may be much smaller than $t$ (e.g., $s$=200 and $t$=300). 
This forms the core of DDIM sampling process:
\begin{equation}\label{eq:ddim-samplingfast}
\tilde{x}_{s}=\frac{\sqrt{\bar{\alpha}_{s}}}{\sqrt{\bar{\alpha}_t}} x_t -
	\left(\frac{\sqrt{\bar{\alpha}_{s}}}{\sqrt{\bar{\alpha}_t}}\sqrt{1-\bar{\alpha}_t}
        -\sqrt{1-\bar{\alpha}_{s}}\right)
        \epsilon_{\theta}^{(t)} .
\end{equation}
In \eqref{eq:ddim-samplingfast}, the sampling step size is controlled by $\bar{\alpha}_t$ and $\bar{\alpha}_s$ only. 
Consequently, we define a $K$-step \textit{sampling trajectory} as the sequence of $\bar{\alpha}$ values 
$\{\bar{\alpha}_1, \bar{\alpha}_2, \cdots, \bar{\alpha}_{K}\}$, 
where each value is a scalar $\bar{\alpha}\in[0,1]$, representing the noise level of a step in the sampling process. 
Following a sampling trajectory $\{\bar{\alpha}_k\}_{k=1}^{K}$, 
diffusion models can convert Gaussian noise to a clean sample. 
Our goal is to find a better trajectory $\{\bar{\alpha}^*_k\}_{k=1}^{K}$ 
that makes the generation result closer to ground-truth distribution $q(x_0)$. 

\section{Related Work}
Since this work focuses on sampling trajectory, we categorize related sampling methods based on trajectory type. 

\subsection{Sampling with Predefined Trajectory}\label{sec:traj-predefined}

Some methods aim to improve sampling quality without model-specific training or modification. 
For this purpose, they increase the discretization accuracy of the differential equations. 
Song \etal\ \cite{song2020score} introduced a predictor-corrector framework to correct errors 
in the evolution of the discretized reverse-time stochastic differential equations (SDE), 
and Lu \etal\ \cite{lu2022dpm} leveraged Taylor expansion to assist in solving the diffusion ordinary differential equations (ODE). 
The former generalizes DDPM and the latter generalizes DDIM. 
Notably, by exploiting the ODE solution, Lu \etal\ derived the DPM-Solver sampler. 
It permits three orders of Taylor expansion, where the order-1 expansion is equivalent to DDIM. 
Recent works DEIS \cite{zhang2022deis}, PNDM \cite{liu2022pndm} and SplitNM \cite{wiz2023splitnm} 
attempted to resolve more precise ODE solution by numerical methods, 
including Runge-Kutta \cite{runge_kutta}, Adams-Bashforth \cite{hochbruck_ostermann_2010} and Strang splitting \cite{condi_stsp}. 
Meanwhile, IIA \cite{iia2024} aimed to improve the accuracy of the integration approximation by minimizing a mean squared error (MSE) function. 
These methods achieved impressive performance. 
However, due to their reliance on predefined trajectories, they primarily focus on optimizing the accuracy of individual sampling steps, 
and overlook the potential of optimizing the entire sampling trajectory. 

\subsection{Sampling with Learning-Based Trajectory}

Watson \etal\ \cite{watson2021l_dynamic_size} suggested selecting the best \textit{K} timesteps to maximize the training objective. 
Jolicoeur \etal\ \cite{dblp_sde} introduced a dynamic step size algorithm by resolving SDE integration. 
Both methods are specific to DDPM models and their generation results are stochastic. 
Kingma \etal\ \cite{nips_vdm} proposed to jointly optimize two networks, $\theta$ and $\eta$, 
where $\theta$ is the diffusion model and $\eta$ is used to learn the sampling trajectory. 
Its number of sampling steps is pre-set to be the same as in the diffusion process 
and therefore, it cannot be applied as off-the-shelf tools. 
Li \etal\ \cite{autodiff} presented AutoDiffusion, which aims to find better trajectories 
by searching through potential timestep sequences and modifying model architectures. 
However, it uses Frechet Inception Distance (FID) as the performance estimation for candidates in the search space, 
making it cumbersome and time-consuming. 

The proposed VRG method belongs to the learning-based category. 
However, different from the aforementioned methods, VRG focuses on the influence of cumulative prediction error in the sampling process. 
By reducing such influence, it is able to optimize sampling trajectory and improve sample quality. 
In addition, VRG is training-free, model-agnostic and its number of sampling steps can be easily configured. 

\section{Method}\label{sec:algo}
In this section we delve deeper into the nature of predicted 
noise $\epsilon_\theta^{(t)}$ in \eqref{eq:ddim-sampling-pred}, 
and then analyze the cumulative prediction error in sampling process. 
Finally, we present VRG, our method for sampling trajectory optimization.

\subsection{Distribution of Prediction Error}\label{sec:algo-pe-distri}
At any timestep $t$, prediction error is the difference between predicted noise $\epsilon_{\theta}^{(t)}$ 
and ground-truth noise $\epsilon^{(t)}$. 
The training loss function $\|\epsilon_{\theta}^{(t)}-\epsilon^{(t)}\|_2^2$ is the $\ell_2$-norm of the prediction error. 
Given a well-trained diffusion model and a large number of samples $x_t$, 
we find that the prediction error approximates a Gaussian distribution with a mean of \textbf{0}:
\begin{equation}\label{eq:train-error}
\begin{split}
    &\epsilon_{\theta}^{(t)}-\epsilon^{(t)} \sim \mathcal{N}(\textbf{0}, \mathbf{\Sigma_{\theta}}) .
\end{split}
\end{equation}
The distribution of prediction error is verified in Sec. \ref{sec:exp-pe-distri}.
We can get the averaged variance of all dimensions in the prediction error: 
\begin{equation}\label{eq:delta-epsilon-t}
\begin{split}
    \Delta_{\epsilon}^{(t)} &= \frac{1}{d} \mathbb{E}_{q(x_t)}\big[\|\epsilon_{\theta}^{(t)}-\epsilon^{(t)}\|_2^2\big] , \\
\end{split}
\end{equation}
where $d$ is the dimensionality of the noise. 
The VRG method disregards individual differences and covariances between dimensions 
in the prediction error, and focuses on the overall impact instead. 
Thus, we use $\Delta_{\epsilon}^{(t)}$ to represent the variance of prediction error of timestep $t$. 
In the rest of the paper, \textit{prediction error} refers to $\Delta_{\epsilon}^{(t)}$ unless stated otherwise. 
\begin{equation}\label{eq:train-error-2}
    \epsilon_{\theta}^{(t)}-\epsilon^{(t)} \sim \mathcal{N}(\textbf{0}, \Delta_{\epsilon}^{(t)} \mathbf{I}) .
\end{equation}

\subsection{Variance of Cumulative Prediction Error}
\label{sec:train-error}
Based on the DDIM equations in \eqref{eq:ddim-sampling} and \eqref{eq:ddim-sampling-pred}, 
the difference between the generated sample and the ground-truth sample is: 
\begin{equation}\label{eq:vubo5}
\begin{split}
\tilde{x}_{t\text{-}1} - x_{t\text{-}1} &= -\frac{\sqrt{1-\bar{\alpha}_t} - \sqrt{\alpha_t-\bar{\alpha}_t}}{\sqrt{\alpha_t}}
        (\epsilon_{\theta}^{(t)}-\epsilon^{(t)})  \\
    &\sim \mathcal{N}\Big(
        \textbf{0}, \frac{\left(\sqrt{1-\bar{\alpha}_t}-\sqrt{\alpha_t-\bar{\alpha}_t}\right)^2}{\alpha_t} \Delta_{\epsilon}^{(t)} \textbf{I}
        \Big) .
\end{split}
\end{equation}
Here, $x_{t-1}$ is the fixed ground truth and $\tilde{x}_{t-1}$ is the predicted value. 
The exact difference between them is intractable due to the unknown nature of $x_{t-1}$. 
Nevertheless, the distribution of their difference is shown in \eqref{eq:vubo5}, 
allowing us to treat each predicted value as a stochastic variable. 
\begin{equation}\label{eq:gen-error}
\begin{split}
\tilde{x}_{t-1} \sim \mathcal{N}\Big(
        x_{t-1}, \frac{\left(\sqrt{1-\bar{\alpha}_t}-\sqrt{\alpha_t-\bar{\alpha}_t}\right)^2}{\alpha_t} \Delta_{\epsilon}^{(t)} \textbf{I}
    \Big) .
\end{split}
\end{equation}
Finally, by the properties of Gaussian distribution and Markov chain, 
we obtain the generated sample $\tilde{x}_0$ (detailed deduction is provided in supplemental material): 
\begin{equation}\label{eq:x-tilde-0}
\begin{split}
    & \tilde{x}_0 \sim \mathcal{N}\Big(x_0,	
        \sum_{t=1}^{T}\big(w(\bar{\alpha}_t, \alpha_t) \Delta_{\epsilon}^{(t)}\big) \mathbf{I}\Big) \\
    & w(\bar{\alpha}_t, \alpha_t) = \frac{\big(\sqrt{1-\bar{\alpha}_t}-\sqrt{\alpha_t-\bar{\alpha}_t}\big)^2}{\bar{\alpha}_t} ,
\end{split}
\end{equation}
where $w(\bar{\alpha}_t, \alpha_t)$ is the weight of the prediction error $\Delta_{\epsilon}^{(t)}$. 
Therefore, the final variance is the sum of weighted prediction errors. 
In \eqref{eq:x-tilde-0}, the mean value is fixed 
but the variance is dynamic and can be reduced by optimizing the sampling trajectory. 
We define the variance magnitude as \textit{variance of cumulative prediction error}. 
If the variance decreases, the cumulative prediction error will be reduced, which potentially improves generation quality. 
However, in the sampling process, the ground truth $\epsilon^{(t)}$ is not accessible, 
making the prediction error $\Delta_{\epsilon}^{(t)}$ unknown. 
To address this, we propose an approach to approximate $\Delta_{\epsilon}^{(t)}$ in the following section. 

\subsection{Mapping between $\bar{\alpha}_t$ and $\Delta_{\epsilon}^{(t)}$}
\label{sec:mapping}
The proposed VRG method uses the prediction error from the training process 
to represent the prediction error in the sampling process. 
In a well-trained model, these two processes exhibit similar prediction error, as verified in Sec. \ref{sec:exp-pred}. 

The diffused sample $x_t$ contains some noise, and the model $\epsilon_{\theta}(x_t, t)$ is used to predict the noise. 
The prediction error depends on the proportion of noise in $x_t$, 
which in turn is determined by the value of $\bar{\alpha}_t$, as shown below: 
\begin{equation}\label{eq:delta-with-x0}
    \Delta_{\epsilon}^{(t)} = \frac{1}{d} \mathbb{E}\big[\|
        \epsilon_{\theta}\big(\underbrace{\sqrt{\bar{\alpha}_t}x_0+\sqrt{1-\bar{\alpha}_t}\epsilon^{(t)}}_{x_t}, t \big) 
        - \epsilon^{(t)}
    \|_2^2\big] ,
\end{equation}
where $\epsilon^{(t)}$ is ground-truth noise, and both instances of $\epsilon^{(t)}$ in \eqref{eq:delta-with-x0} are the same. 
When $\bar{\alpha}_t$ is small, the input $x_t$ has a higher proportion of noise $\epsilon^{(t)}$,
making the noise prediction easier and the prediction error smaller. 
Conversely, as $\bar{\alpha}_t$ increases, noise prediction becomes more challenging, and prediction error tends to rise. 
This is evaluated in Sec. \ref{sec:exp-pred}. 

\begin{algorithm}[t]
\caption{$\bar{\alpha}_t$ and $\Delta_{\epsilon}^{(t)}$ calculation}
\label{alg:mse}
\raggedright
\textbf{Input:} Diffusion model $\epsilon_{\theta}$, dataset $D$.\\
\textbf{Output:} $\{\bar{\alpha}_t\}_{t=1}^T$, $\{\Delta_{\epsilon}^{(t)}\}_{t=1}^T$.
\begin{algorithmic}[1]
\For {$t=1$ to $T$}
    \State Calculate $\bar{\alpha}_t$ based on predefined noise schedule
    \For{each sample $x_i$ in $D$}
        \State Generate $\epsilon^{(t)}\sim\mathcal{N}(0,\mathbf{I})$ 
        \State $\epsilon_{\theta}^{(t)}\leftarrow\epsilon_{\theta}(\sqrt{\bar{\alpha}_t}\ x_i + \sqrt{1-\bar{\alpha}_t}\ \epsilon^{(t)}, t)$
        \State $\delta_i \leftarrow \frac{1}{d} \|\epsilon_{\theta}^{(t)}-\epsilon^{(t)}\|_2^2$, \quad $d$ is dimensionality
    \EndFor
    \State $\Delta_{\epsilon}^{(t)} \leftarrow \frac{1}{\ |D|\ }\sum_{i=1}^{n}\delta_i $
\EndFor
\end{algorithmic}
\end{algorithm}
To model the relationship between $\bar{\alpha}_t$ and $\Delta_\epsilon^{(t)}$, 
we define a continuous function $f_\Delta: [0, 1] \rightarrow \mathbb{R}^+$, 
such that $f_\Delta(\bar{\alpha}_t)=\Delta_{\epsilon}^{(t)}$. 
Although the precise definition of $f_\Delta(\cdot)$ is unknown, its input and output are both available. 
The values of $\bar{\alpha}_t$ can be easily obtained by the predefined noise schedule, 
and $\Delta_{\epsilon}^{(t)}$ can be calculated using \eqref{eq:delta-with-x0}. 
Algorithm \ref{alg:mse} provides a detailed procedure for collecting the input-output pairs 
$(\bar{\alpha}_t, \Delta_{\epsilon}^{(t)})$. 
Once these pairs are collected, we can approximate $f_\Delta(\cdot)$ by linear interpolation. 


\subsection{Variance-Reduction Guidance}

\begin{figure}[t]
\centering
{\includegraphics[width=0.45\textwidth]{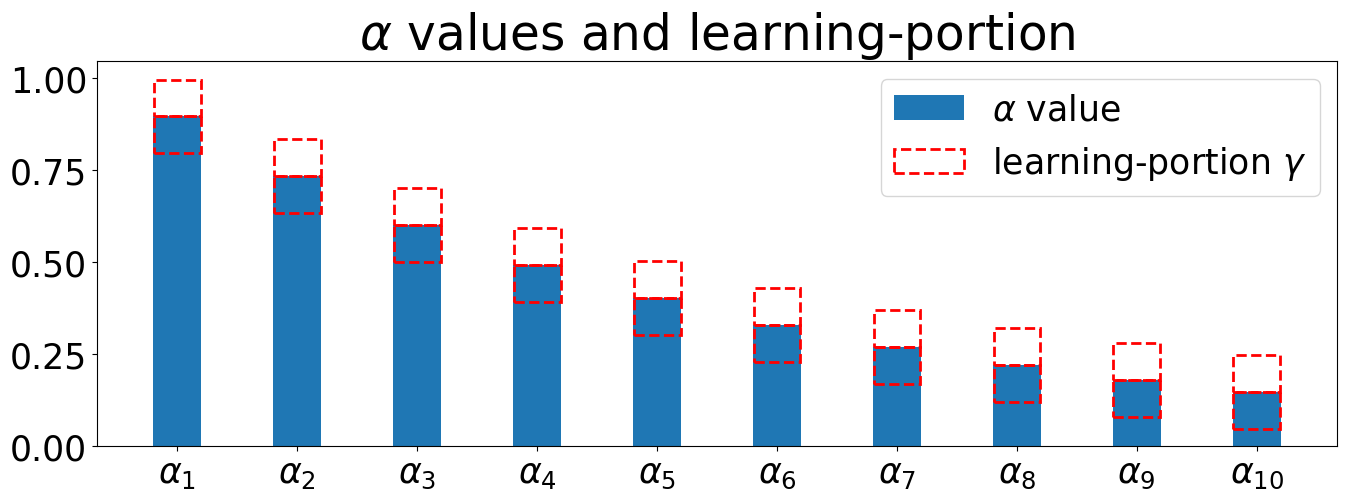}}
\caption{10-step trajectory $\{\alpha_k\}_{k=1}^{10}$ and learning-portion $\gamma$. } 
\label{fig:alphabar10-tu-lp}
\vspace{-0.4cm}
\end{figure}

The VRG method is designed to search for a better sampling trajectory 
based on a predefined trajectory $\{\bar{\alpha}_k\}_{k=1}^K$. 
To ease the optimization, the trajectory is rewritten using $\alpha$ values $\{\alpha_k\}_{k=1}^K$, 
where $\alpha_k=\bar{\alpha}_k/\bar{\alpha}_{k-1}$ assuming $\bar{\alpha}_0=1$. 
It is worth noting that $\{\bar{\alpha}_k\}_{k=1}^K$ and $\{\alpha_k\}_{k=1}^K$ 
are equivalent representations of the sampling trajectory. 
The objective function of VRG is given in \eqref{eq:local-opt-loss1}. 
\begin{equation}\label{eq:local-opt-loss1}
\begin{split}
&\min_{\{\alpha_k^*\}_{k=1}^K} \underbrace{\sum_{k=1}^{K}w(\bar{\alpha}_k^*, \alpha_k^*)f_{\Delta}(\bar{\alpha}_k^*)}_{\text{cumulative prediction error}}
    +\underbrace{\vphantom{\sum_{k=1}^{K}00} \lambda ||\bar{\alpha}_K^* - \bar{\alpha}_K||_2^2}_{\text{regularizer}} ,\\
    &\begin{split}
    \text{s.t.} &\ \alpha_k^* \in (0, 1) , \ |\alpha_k^* - \alpha_k| \le \gamma \quad and \quad \bar{\alpha}_k^* = \prod_{s=1}^k \alpha_s^* ,\\
    \end{split}
\end{split}
\end{equation}
where $\{\alpha_k^*\}_{k=1}^K$ is the new trajectory, 
$\lambda$ controls the weight of the regularizer, 
and $\gamma$ constrains the difference between new and old trajectories. 
We denote $\gamma$ as \textit{learning-portion}. 
Fig. \ref{fig:alphabar10-tu-lp} illustrates $\gamma$ with a 10-step trajectory $\{\alpha_k\}_{k=1}^{10}$. 


In \eqref{eq:local-opt-loss1}, the \textit{cumulative prediction error} part applies to the new trajectory, 
where $\alpha_k^*$ is restricted within a range of $\alpha_k$ defined by $\gamma$. 
Meanwhile, the \textit{regularizer} part keeps the final cumulative product $\bar{\alpha}_K^*$ close to $\bar{\alpha}_K$. 
It ensures that the new trajectory starts from similar noise level as the old trajectory. 
We constructed a neural network using the projected gradient descent (PGD) 
technique \cite{pgd_2018_adv_attack} to model the objective function. 
Additional details are provided in the supplemental material. 
Given an existing trajectory, searching for a new trajectory is a one-time process that completes within one minute. 

\section{Experiments}\label{sec:exp}
Experiments are conducted on multiple datasets, pre-trained models, and baseline methods. 
The datasets are CIFAR10 \cite{url_dset_cifar10}, LSUN-Bedroom \cite{url_dset_lsun-bedroom}, 
ImageNet \cite{deng2009imagenet}, CelebA \cite{url_dset_celeba} 
and a synthetic image set generated by Stable Diffusion \cite{stable_diffusion}. 
CIFAR10 and LSUN-Bedroom use pre-trained model checkpoints from Patrick Esser \cite{github_pesser}, 
ImageNet uses checkpoints from OpenAI \cite{github_openai_g_diff}, 
and CelebA model is trained by ourselves following \cite{iclr_ddim}. 
Furthermore, seven baseline methods are DDIM, DPM-Solver \cite{lu2022dpm}, DEIS \cite{zhang2022deis}, 
PNDM \cite{liu2022pndm}, SplitNM \cite{wiz2023splitnm}, Stable Diffusion \cite{stable_diffusion}, and AutoDiffusion \cite{autodiff}. 
For each baseline, we track the vanilla sampling trajectory, optimize it by VRG, 
and then embed the new trajectory back into its sampling process. 
In this way, we generate samples by both trajectories while keep other settings consistent. 
For quality evaluation, we use the FID score \cite{fid_gan}, 
where lower score implies better quality.

\subsection{Comparison with DDIM}
\begin{figure}[t]  
  \centering
  \includegraphics[width=0.95\linewidth]{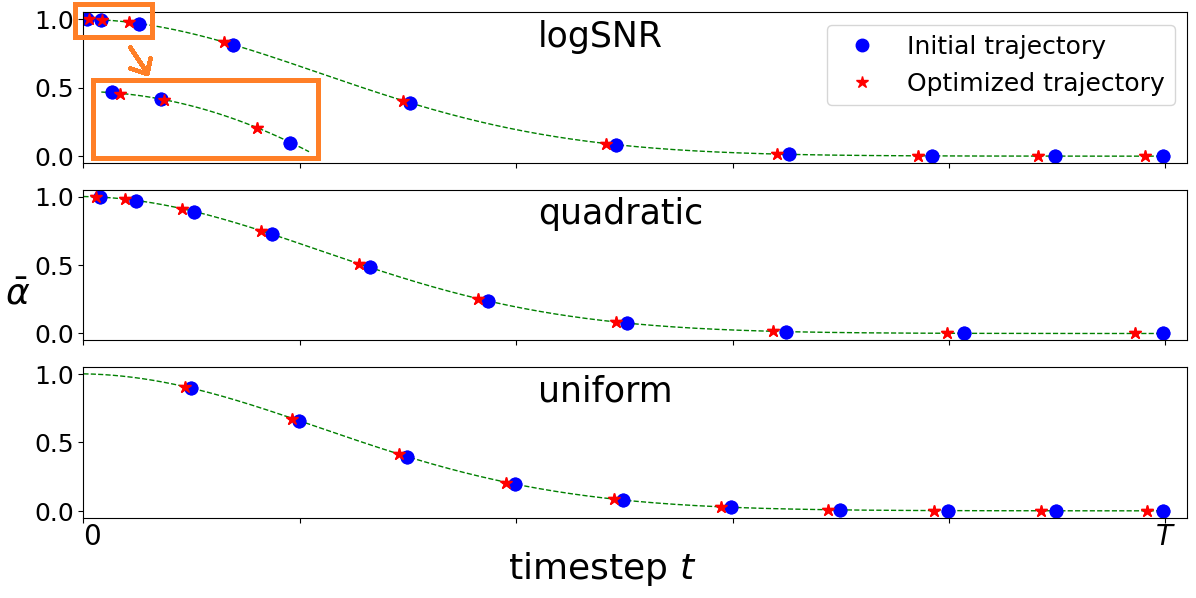}
  \vspace{-0.2cm}
  \caption{Sampling trajectory optimization on different trajectories: $logSNR$, $quadratic$ and $uniform$. 
          }
  \label{fig:abc-01-s10-all3-zzz}
\end{figure}
\begin{figure}[t] 
    \begin{center}
    \includegraphics[width=0.9\linewidth]{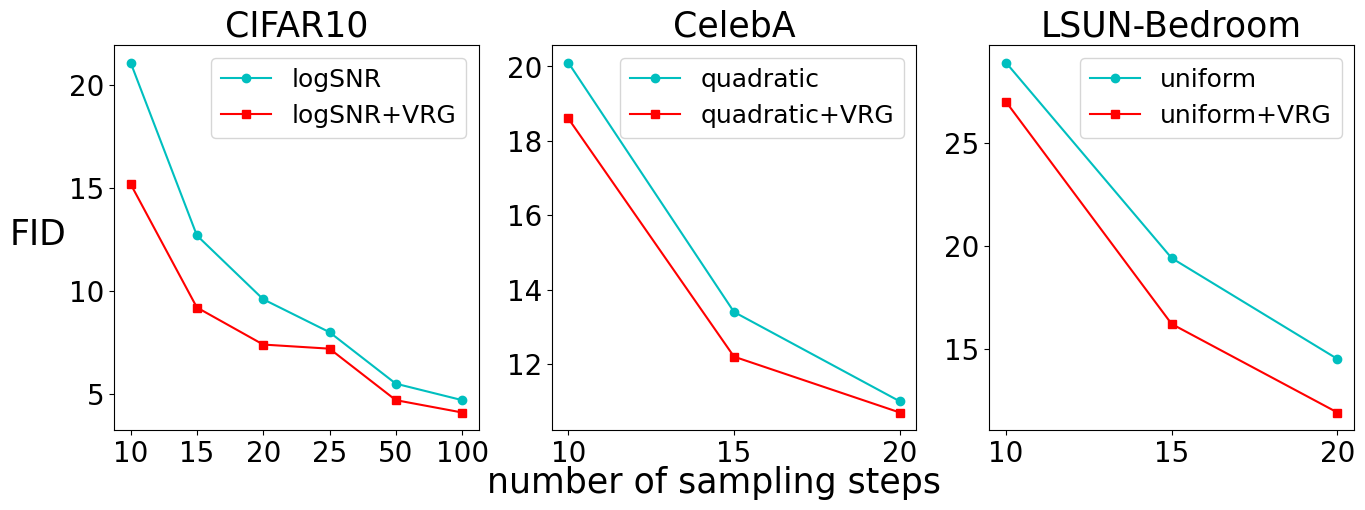}
    \end{center}
    \vspace{-0.3cm}
    \caption{FID$\downarrow$ comparison with DDIM on different trajectories. }
    \label{fig:fid-ddim-cifar10}
\end{figure}
\begin{table}[!ht]
\centering
\caption{FID$\downarrow$ comparison with DPM-Solver on CIFAR10. }
\label{tab:fid-dpm-cifar10}
\begin{tabular}{ c l r r r r r r }
\toprule
& & \multicolumn{6}{c}{Sampling steps} \\
\cmidrule{3-8}
Order & Trajectory & 10 & 15 & 20 & 25 & 50 & 100 \\
\midrule
\multirow{6}{*}{2}
& logSNR        & 7.5 & 4.0 & 4.1 & 3.9 & 4.1 & 4.0 \\
& logSNR+VRG	& \textbf{3.8} & \textbf{3.5} & \textbf{3.5} & \textbf{3.4} & \textbf{3.6} & \textbf{3.5} \\
& quadratic     & 39.1 & 8.1 & 4.3 & 3.7 & 4.0 & 4.1 \\
& quadratic+VRG	&  \textbf{3.7} & \textbf{3.5} & \textbf{4.1} & \textbf{3.5} & \textbf{3.7} & \textbf{3.5} \\
& uniform       & 10.2 & 7.4 & 5.2 & 5.0 & 3.8 & 3.9 \\
& uniform+VRG	&  \textbf{4.6} & \textbf{3.2} & \textbf{4.9} & \textbf{4.3} & \textbf{3.6} & \textbf{3.7} \\
\hline
\multirow{6}{*}{3}
& logSNR        & 6.4  & 3.7 & 4.3 & 3.7 & 4.1 & 4.0 \\
& logSNR+VRG	& \textbf{5.7} & \textbf{3.4} & \textbf{4.2} & \textbf{3.2} & \textbf{3.7} & \textbf{3.5} \\
& quadratic     & 168  & 16.8 & 6.4 & 4.1 & 4.1 & 4.0 \\
& quadratic+VRG	&  \textbf{42.9} &  \textbf{4.1} & \textbf{5.2} & \textbf{3.6} & \textbf{3.6} & \textbf{3.5} \\
& uniform       & 12.6 & 7.1 & 5.2 & 5.0 & 3.8 & 3.5 \\
& uniform+VRG	&  \textbf{5.8} & \textbf{6.4} & \textbf{4.9} & \textbf{4.3} & \textbf{3.5} & \textbf{3.2} \\
\bottomrule
\end{tabular}
\vspace{-0.2cm}
\end{table}
To compare with DDIM, we tested three datasets: CIFAR10, LSUN-Bedroom and CelebA. 
Following DPM-Solver, we adopt three strategies for the predefined trajectory: 
\textit{logSNR}, \textit{quadratic} and \textit{uniform}. 
A visual comparison of them is shown in Fig. \ref{fig:abc-01-s10-all3-zzz}, 
indicating that VRG is self-adaptive and can optimize different trajectories. 
For quantitative comparison, Fig.~\ref{fig:fid-ddim-cifar10} depicts the FID scores, 
where VRG consistently improves generation quality regardless of dataset and trajectory. 
Additional comparison figures are included in the supplemental material. 

\subsection{Comparison with DPM-Solver}
The core idea of DPM-Solver lies in the $order$ of its Taylor expansion. 
Specifically, order 1 is equivalent to DDIM, which we have already compared. 
In this section we focus on orders 2 and 3. 
The pre-trained model checkpoint \cite{github_pesser} is discrete-time model with linear-scheduled noise. 
The quantitative comparison is provided in Table \ref{tab:fid-dpm-cifar10}. 
The FID values reported in this table may differ from those in \cite{lu2022dpm} 
because we were not able to find the exact model checkpoint used in that paper. 
For visual comparison, Fig. \ref{fig:compare-ab-img}(b) showcases the generation results 
on LSUN-Bedroom dataset, demonstrating the improvements in image quality and fidelity. 

\subsection{Comparison with DEIS, PNDM, SplitNM and Others}
\begin{figure}[t]
\centering
  \subcaptionbox{DEIS}
  {\includegraphics[width=.32\linewidth]{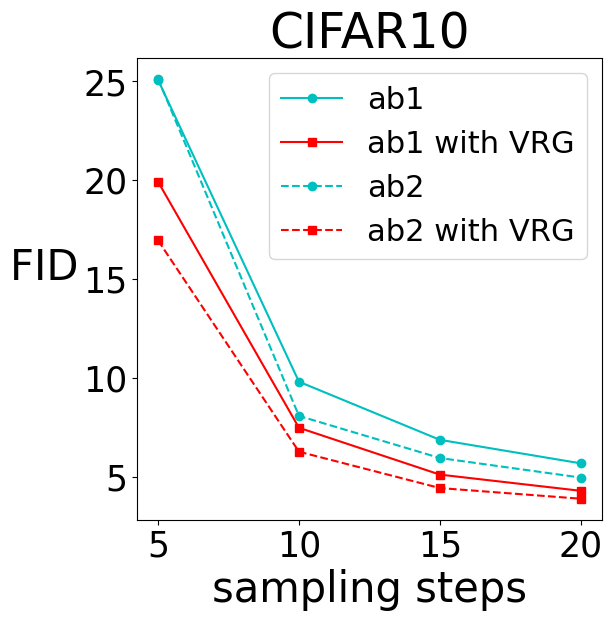}}
  \subcaptionbox{PNDM}
  {\includegraphics[width=.32\linewidth]{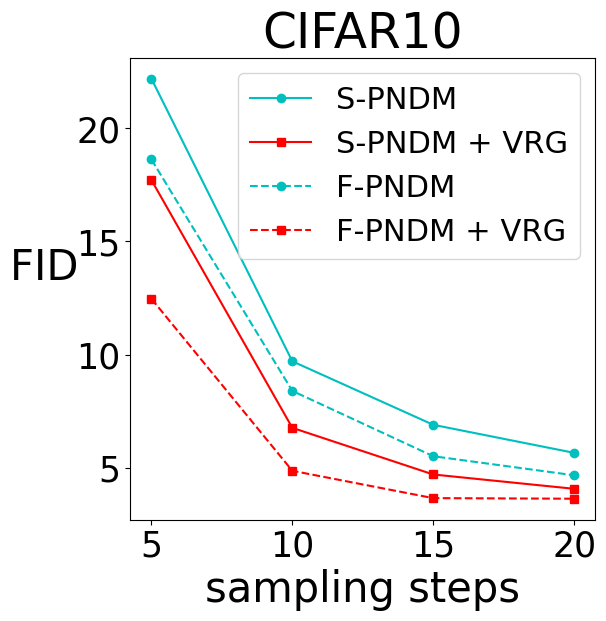}}
  \subcaptionbox{SplitNM}
  {\includegraphics[width=.32\linewidth]{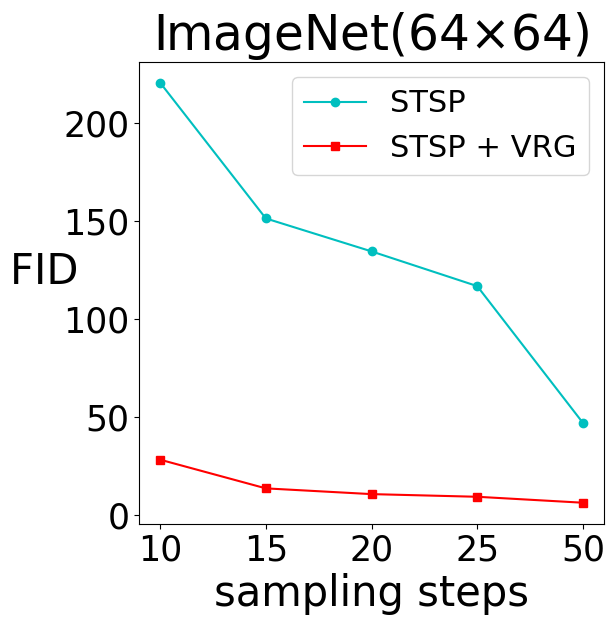}}
\caption{FID$\downarrow$ comparison on different baselines. 
         (a)  ``ab1'' and ``ab2'' denote $ab\_order$=1 and $ab\_order$=2 respectively.
        }
\label{fig:deis-pndm-splitnm}
\vspace{-0.2cm}
\end{figure}

Diffusion Exponential Integrator Sampler (DEIS) \cite{zhang2022deis} is designed 
for discretizing ODEs by the Exponential Integrator technique \cite{hochbruck_ostermann_2010}. 
DEIS has two important hyperparameters, $ts\_order$ and $ab\_order$, 
where the former is the order of timestep controlling the sampling trajectory, 
and the latter denotes the order of the Adams-Bashforth method used for integration. 
When running DEIS, we keep $ts\_order$=1 and vary $ab\_order$ and the number of sampling steps. 
As shown in Fig. \ref{fig:deis-pndm-splitnm}(a), 
VRG can improve the trajectories for all the cases and enhance their generation quality. 

Pseudo Numerical methods for Diffusion Models (PNDM) \cite{liu2022pndm} 
treat the sampling process as solving differential equations on manifolds. 
PNDM has two variants: S-PNDM and F-PNDM. 
The former combines two second-order numerical methods to form a pseudo numerical method, 
and the latter leverages four steps of other methods to make a pseudo step. 
Fig. \ref{fig:deis-pndm-splitnm}(b) depicts the FID comparison for both variants, 
highlighting the superior performance of VRG. 

Guided sampling with Splitting Numerical Methods (SplitNM) \cite{wiz2023splitnm} 
is designed for conditional generation. 
It leverages the high-order Strang splitting \cite{condi_stsp} method (STSP) for guided sampling 
and generates high-quality images with fewer steps compared to traditional methods. 
Its experiments are conducted on ImageNet \cite{deng2009imagenet} 
with pre-trained models from OpenAI \cite{github_openai_g_diff}. 
We use the first 40 categories from ImageNet as ground truth, containing 52K images. 
As shown in Fig. \ref{fig:deis-pndm-splitnm}(c), VRG significantly reduces FID scores, 
showing its effectiveness in improving the sampling trajectory. 

AutoDiffusion \cite{autodiff} jointly searches the timesteps and model architectures in the search progress. 
It relies on the FID score for performance estimation, making it computationally expensive and time-consuming. 
To evaluate our approach, we adopted the sampling trajectories reported in its appendix and applied our VRG method.
Experimental results demonstrate that VRG performs comparably to AutoDiffusion.

\begin{figure}[t]
\centering
  {\includegraphics[width=.49\linewidth]{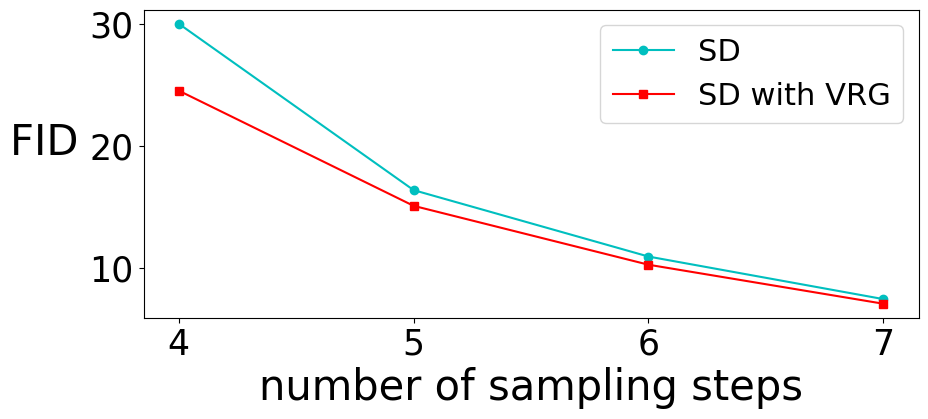}}
  {\includegraphics[width=.49\linewidth]{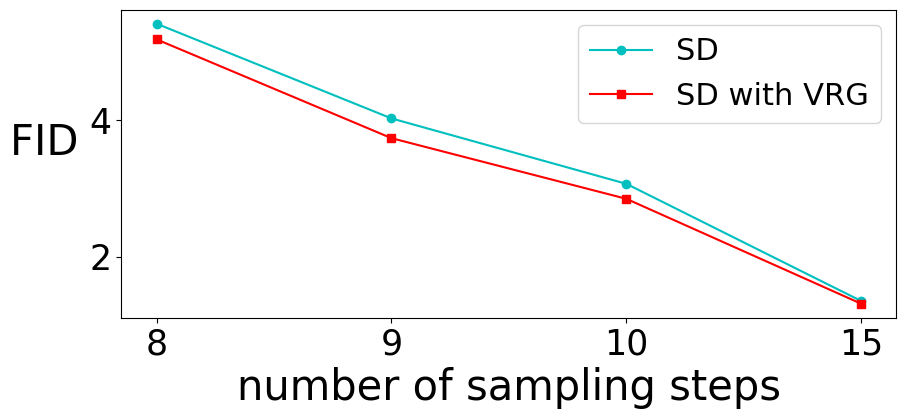}}
\caption{FID$\downarrow$ comparison on Stable Diffusion. 
        }
\label{fig:fid-sd}
\vspace{-0.3cm}
\end{figure}
\begin{figure}[!ht]
\centering
  \subcaptionbox*{3 steps} {\includegraphics[width=.19\linewidth]{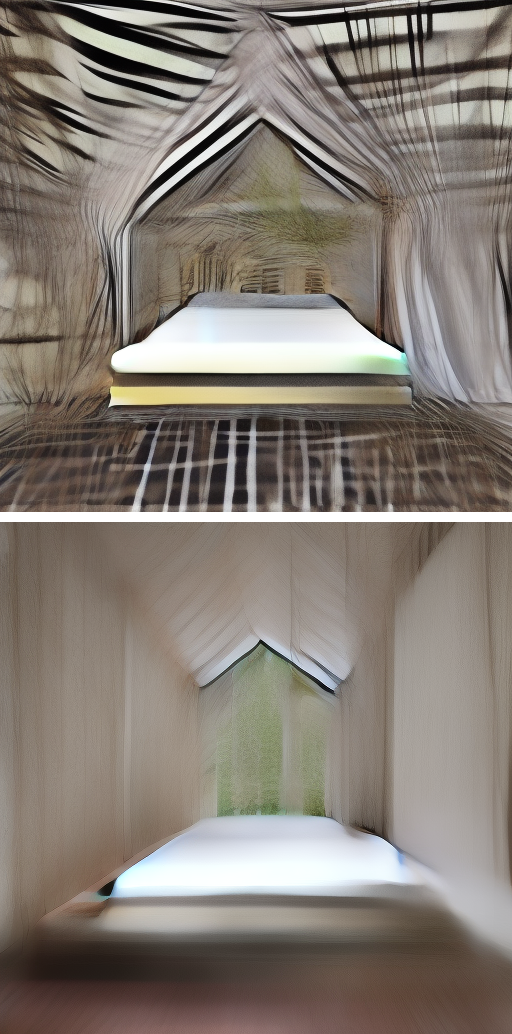}}
  \subcaptionbox*{4 steps} {\includegraphics[width=.19\linewidth]{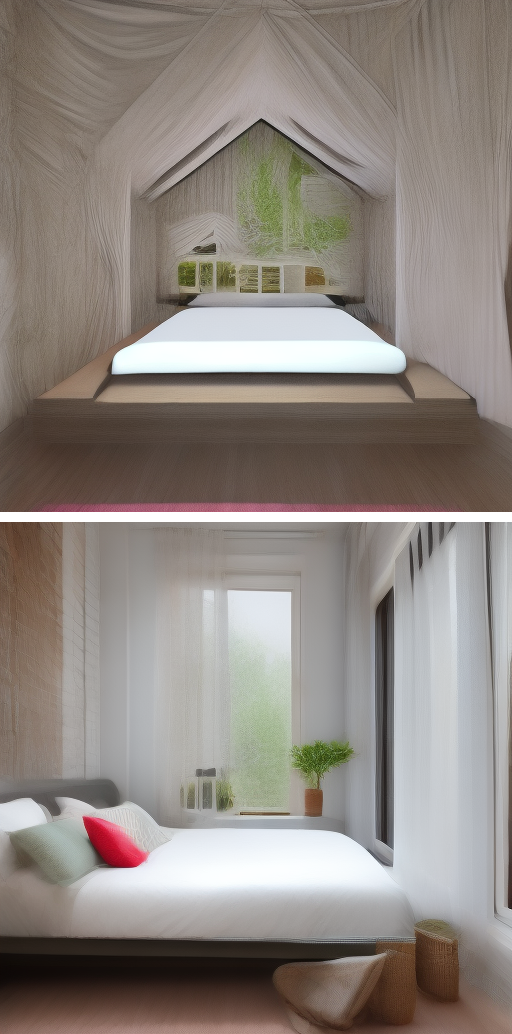}}
  \subcaptionbox*{5 steps} {\includegraphics[width=.19\linewidth]{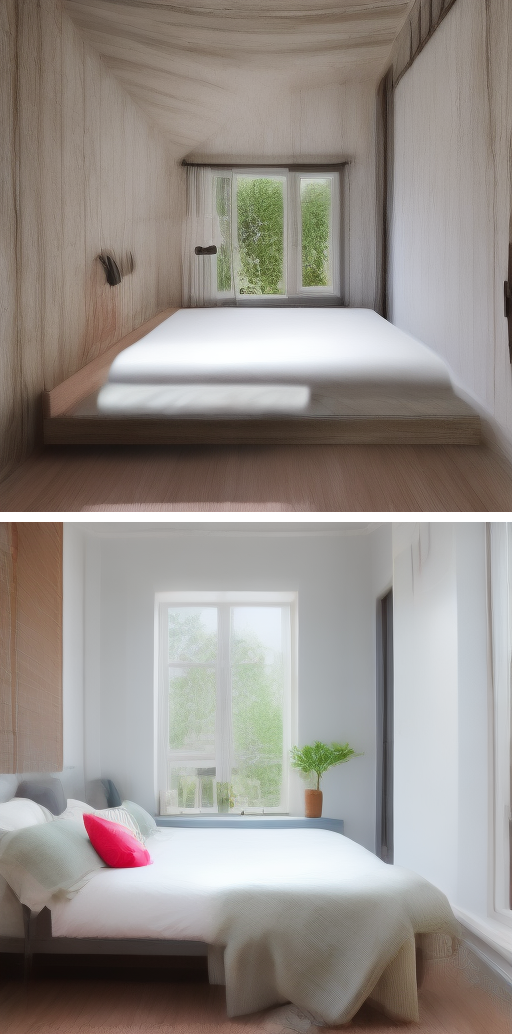}}
  \subcaptionbox*{7 steps} {\includegraphics[width=.19\linewidth]{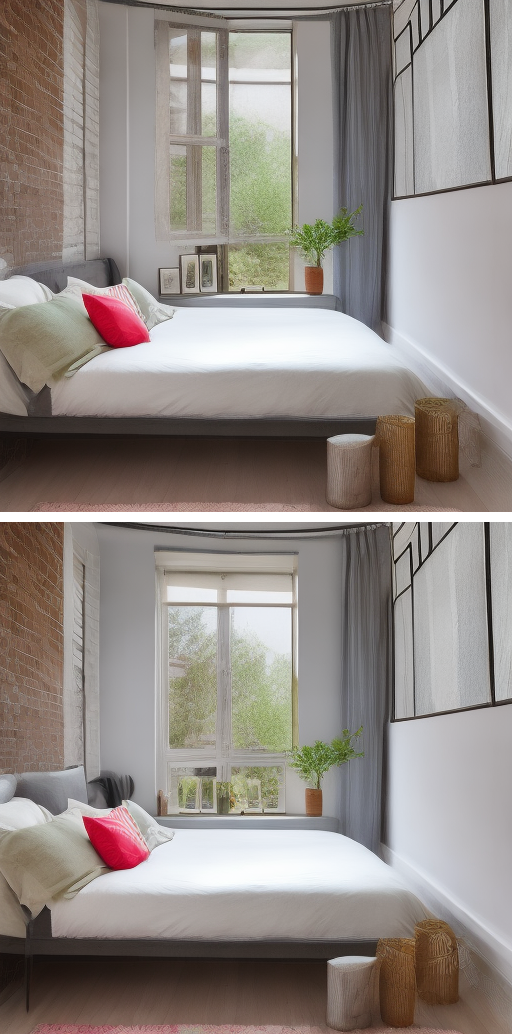}}
  \subcaptionbox*{10 steps}{\includegraphics[width=.19\linewidth]{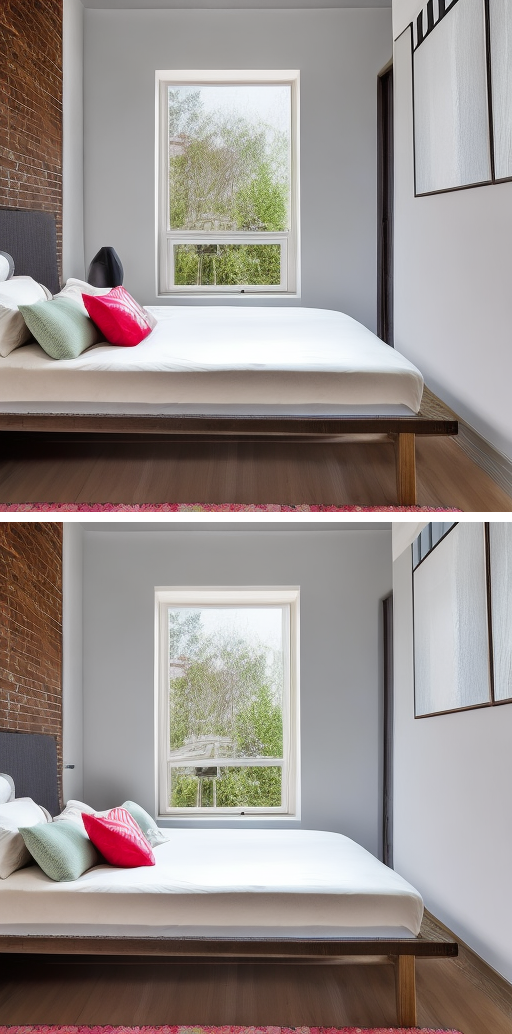}}
\caption{Stable Diffusion: generated images by DDIM sampler (\textit{top}) and DDIM+VRG sampler (\textit{bottom}). 
        }
\label{fig:img-compare-sd}
\vspace{-0.2cm}
\end{figure}
\begin{figure}[!ht]
\centering
  \subcaptionbox{Cumulative prediction error}
    {\includegraphics[width=.495\linewidth]{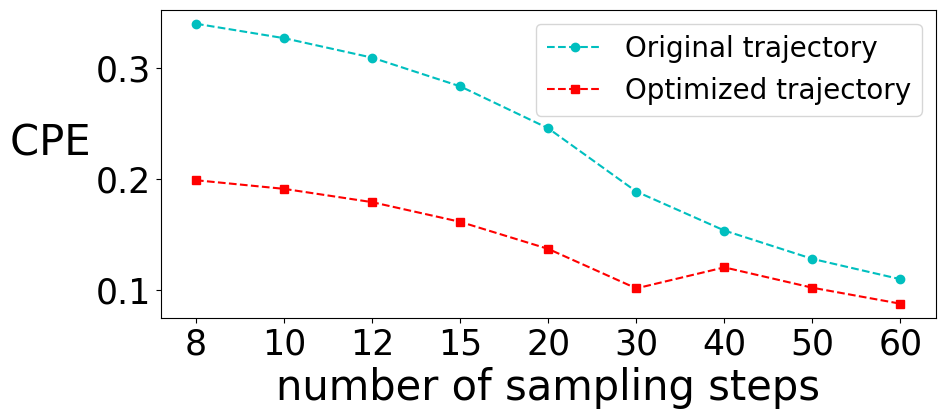}}
  \subcaptionbox{FID score}
    {\includegraphics[width=.48\linewidth]{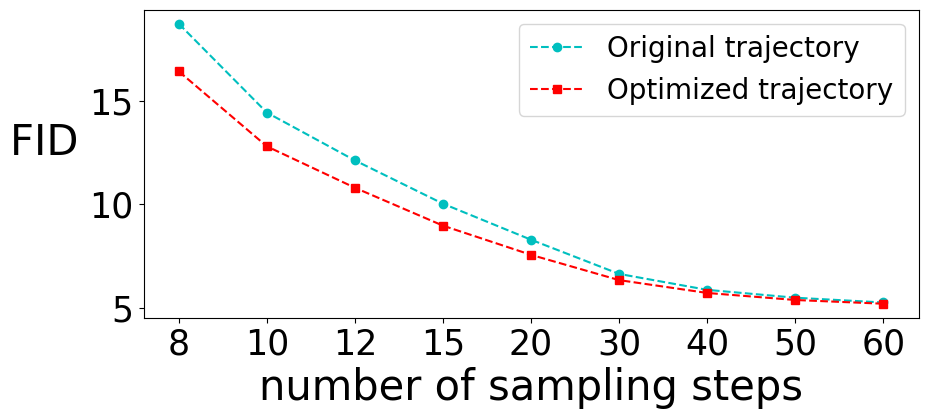}}
\caption{Comparison of cumulative prediction error (CPE) and FID between original 
         and optimized trajectories on CIFAR10. 
        }
\label{fig:cpe_fid_compare}
\vspace{-0.4cm}
\end{figure}
\subsection{Comparison with Stable Diffusion}
Stable Diffusion (SD) \cite{stable_diffusion} is a high-resolution image synthesis model based on latent diffusion models. 
The VRG method can be applied to these latent diffusion models and enhance the generation quality. 
Fig. \ref{fig:fid-sd} depicts a quantitative comparison by FID, 
and Fig. \ref{fig:img-compare-sd} presents a qualitative comparison through visual examples. 
To ensure a fair comparison, we use the same text prompt for ground-truth, 
baseline, and optimized images, all rendered at a resolution of 512$\times$512. 
The ground truth consists of 50K images generated by SD with 50 steps. 
The baselines are generated with fewer steps using vanilla trajectories, 
while the optimized images are generated with the same step counts as baselines, but by optimized trajectories. 

\subsection{Cumulative Prediction Error as an Indicator}
In Fig. \ref{fig:cpe_fid_compare}, the optimized trajectory not only results in reduced FID scores but also lower cumulative prediction error (CPE). 
As the number of sampling steps increases, both the CPE and the FID score demonstrate a consistent downward trend, 
demonstrating that the model's prediction results are becoming more accurate and the generated images are achieving higher fidelity. 
Fig. \ref{fig:cpe_fid_compare} underscores the effectiveness of the VRG method in optimizing sampling trajectory and boosting overall generation quality.

\subsection{Distribution of Prediction Error}
\label{sec:exp-pe-distri}
\begin{figure}[t]
\centering
  \includegraphics[width=.48\linewidth]{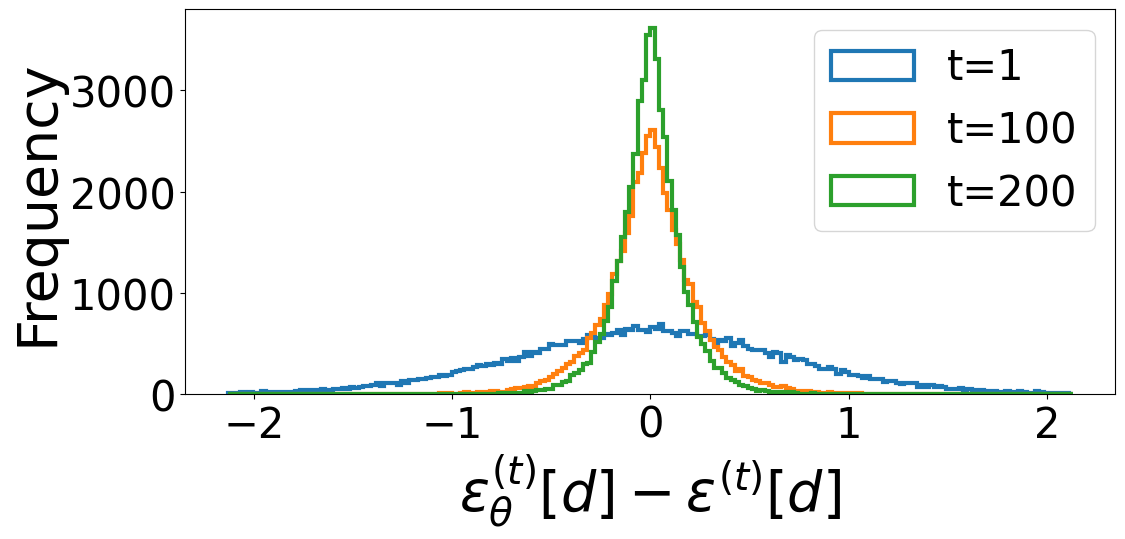}  
  \includegraphics[width=.48\linewidth]{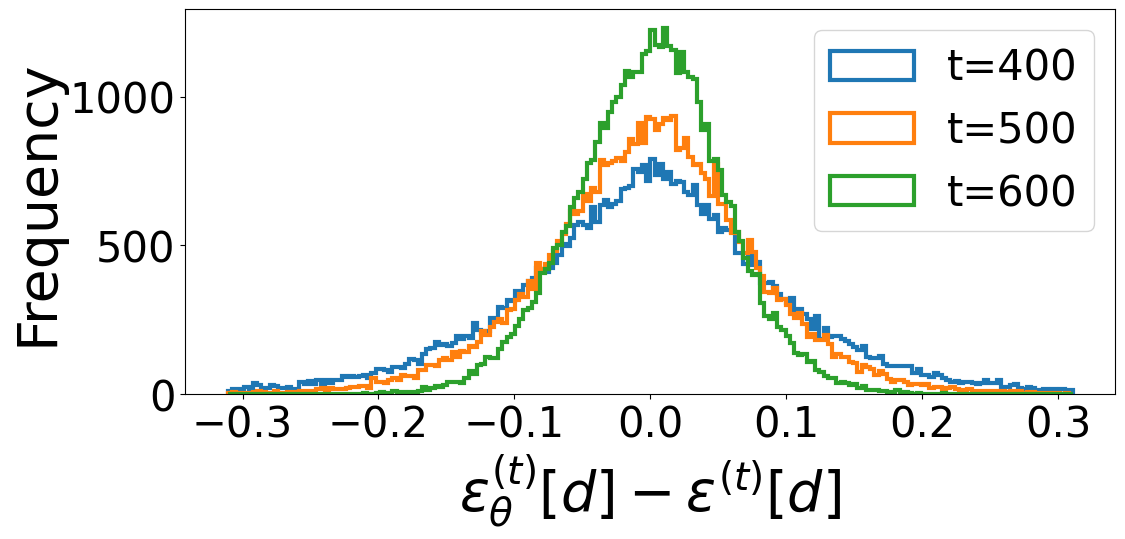}
  \vspace{-0.15cm}
\caption{Difference of predicted noise $\epsilon_{\theta}^{(t)}$ 
         and ground-truth noise $\epsilon^{(t)}$. 
         $t$ is timestep, and $d$ is randomly selected dimension.
        }
\label{fig:predict-error-distri}
\vspace{-0.3cm}
\end{figure}
We evaluate the distribution of prediction error by pre-trained diffusion model \cite{github_pesser} 
and 50K training images from CIFAR10. 
The experiment is performed separately on different timesteps. 
Since CIFAR10 images have dimensions of 3$\times$32$\times$32, the predicted noise $\epsilon_\theta^{(t)}$ 
and ground-truth noise $\epsilon^{(t)}$ each have 3072 dimensions. 
We randomly select one dimension, indexed as $d$, and plot its prediction error distribution. 
The distribution is divided into 200 bins for frequency counting. 
As shown in Fig. \ref{fig:predict-error-distri}, 
The distribution of prediction error closely resembles a Gaussian distribution with a mean value of 0, 
which validates the discussion in Sec.~\ref{sec:algo-pe-distri}.

\subsection{Mapping between $\bar{\alpha}_t$ and $\Delta_{\epsilon}^{(t)}$}
\label{sec:exp-pred}

\begin{figure}[t]
  \begin{center}
  \subcaptionbox{CIFAR-10 (32$\times$32)}
    {\includegraphics[width=.49\linewidth]{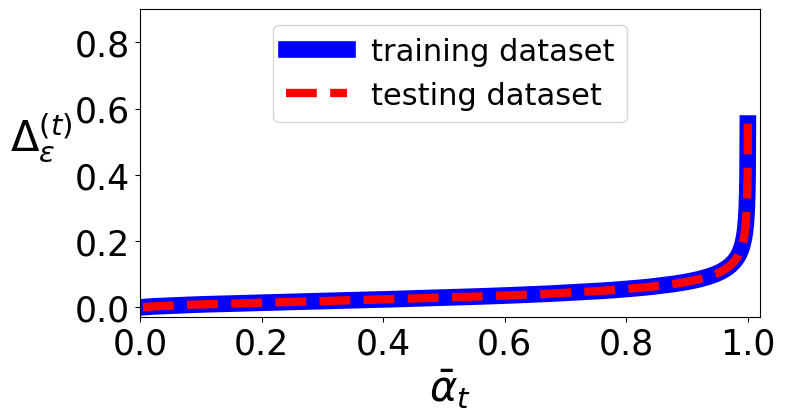}}
  \subcaptionbox{LSUN-Bedroom (256$\times$256)}
    {\includegraphics[width=.49\linewidth]{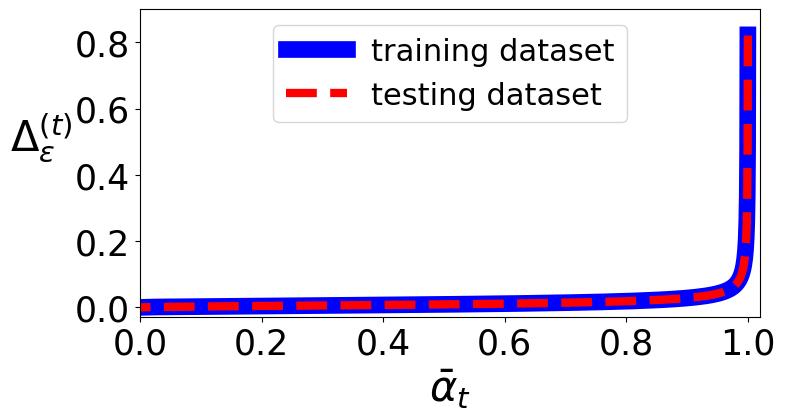}}
  \end{center}
  \vspace{-0.3cm}
  \caption{$\Delta_{\epsilon}^{(t)}\ w.r.t.\ \bar\alpha$ in CIFAR-10 (\textit{left}) and Bedroom (\textit{right}).}
  \label{fig:mse-bar}
\vspace{-0.3cm}
\end{figure}

Through experiments on pre-trained diffusion models, we observe that the values of 
$\Delta_{\epsilon}^{(t)}$ exhibit similar trends in both the training and testing data, 
which are visualized in Fig. \ref{fig:mse-bar}. 
Hence, the function $f_{\Delta}(\cdot)$ will have similar curves in both the training and sampling processes. 
Therefore, given a well-trained model, it is feasible to use prediction error of ground-truth data 
to simulate the prediction error of generated data. 

\subsection{Ablation Study}
\begin{figure}[t]
\centering
  {\includegraphics[width=.49\linewidth]{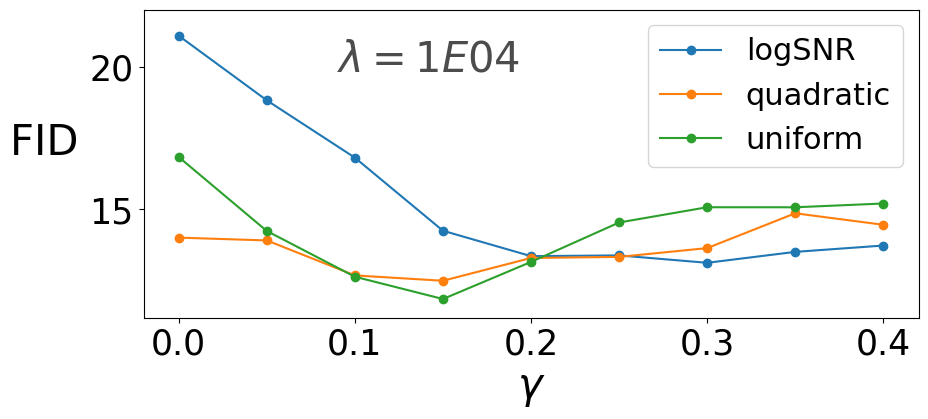}}
  {\includegraphics[width=.49\linewidth]{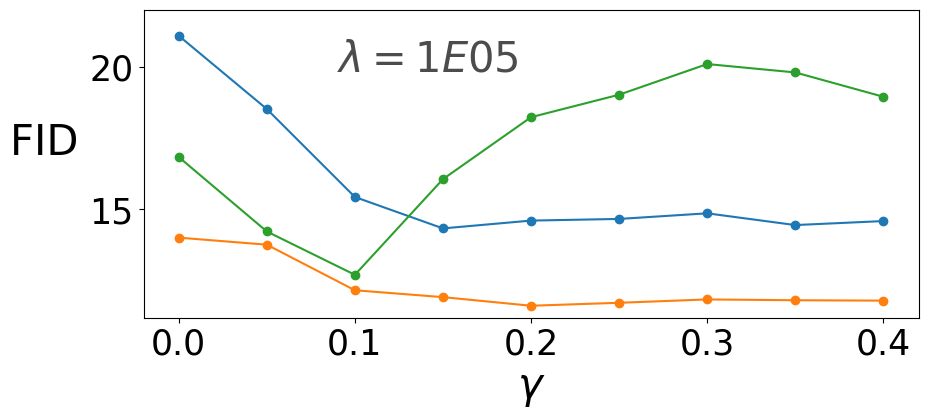}}
\caption{Ablation study on DDIM+VRG with 10-step sampling. 
         It involves different options of $\lambda$, $\gamma$ and trajectories.}
\label{fig:ablation}
\vspace{-0.4cm}
\end{figure}

Based on \eqref{eq:local-opt-loss1}, an ablation study is conducted 
on different values of learning portion $\gamma$ and regularizer weight $\lambda$. 
It is performed on CIFAR10 with a pre-trained model \cite{github_pesser}.
Following DPM-Solver, we make three predefined trajectories by $logSNR$, $quadratic$ and $uniform$. 
In Fig. \ref{fig:ablation}, $\gamma$=0 indicates the original trajectory. 
The FID curves show that a larger $\gamma$ does not necessarily result in better performance. 
If number of steps $K\le10$, $\gamma$ can be 0.1 or 0.05. 
When $K>10$, it is recommended to use a $\gamma$ value of 0.01 or smaller. 

\section{Conclusion}
This paper proposes a novel perspective on the sampling process of diffusion models. 
It derives a metric for the cumulative prediction error and 
introduces the VRG method to optimize the sampling trajectory. 
Comprehensive experiments are conducted on seven baselines across five datasets, 
all of which demonstrate the superiority of the VRG method. 

\section*{Acknowledgment}
This research is supported by the National Research Foundation, Singapore 
and Infocomm Media Development Authority under its Trust Tech Funding Initiative. 
Any opinions, findings and conclusions or recommendations expressed in this material are 
those of the author(s) and do not reflect the views of National Research Foundation, Singapore 
and Infocomm Media Development Authority.

\bibliographystyle{IEEEbib}
\bibliography{icme2025}

\end{document}